\documentclass[twocolumn, 12pt]{article}

\usepackage{titling}
\title{\vspace{-1cm}\bfseries{\Large{Synaptic metaplasticity in binarized neural networks}} \\\vspace*{-0.5cm}}

\author{\normalsize {Axel Laborieux\thanks{Centre de Nanosciences et de Nanotechnologies, Universit\'e Paris-Saclay}, Maxence Ernoult \thanks{Mila, Universit\'e de Montr\'eal}, Tifenn Hirtzlin\textsuperscript{1}, Damien Querlioz\textsuperscript{1}}}

\thanksmarkseries{arabic}

\date{}
\usepackage[T1]{fontenc}
\usepackage[utf8]{inputenc} 
\usepackage[english]{babel}
\usepackage{amssymb,amsthm,amsfonts,amscd}
\usepackage{amsmath}
\usepackage{mathtools}   
\usepackage{fancyhdr}    
\usepackage{makeidx} 
\usepackage{graphicx}
\usepackage{algorithm}
\usepackage{algorithmic}
\usepackage{array}
\usepackage{caption}
\usepackage{lipsum}
\graphicspath{{.},{img/}}

\makeindex
\usepackage{geometry}
\begin{document}
\maketitle
\vspace*{-2.5cm}
\subsection*{Summary}  
\vspace*{-0.2cm}
Unlike the brain, artificial neural networks, including state-of-the-art deep neural networks for computer vision, are subject to ``catastrophic forgetting'' \cite{french}: they rapidly forget the previous task when trained on a new one.
Neuroscience suggests that biological synapses avoid this issue through the process of synaptic consolidation and \emph{metaplasticity}: the plasticity itself changes upon repeated synaptic events \cite{fusi, abraham}.
In this work, we show that this concept of metaplasticity can  be transferred to a particular type of deep neural networks,  binarized neural networks (BNNs) \cite{courbariaux}, to reduce catastrophic forgetting. 
BNNs were initially developed to allow low-energy consumption implementation of neural networks. 
In these networks, synaptic weights and activations are constrained to $\{-1, +1\}$ and training is performed using \emph{hidden} real-valued weights which are discarded at test time.
Our first contribution is to draw a parallel between the metaplastic states of \cite{fusi} and the hidden weights inherent to BNNs.
Based on this insight, we propose a simple synaptic consolidation strategy for the hidden weight.
We justify it using a tractable binary optimization problem, and we
show that our strategy performs almost as well as mainstream machine learning approaches to mitigate catastrophic forgetting, which minimize  task-specific loss functions \cite{kirkpatrick}, on the task of learning  pixel-permuted versions of the MNIST digit dataset sequentially. 
Moreover, unlike these techniques, our approach does not require task boundaries, thereby allowing us to explore a new setting where the network learns from a stream of data.
When trained on data streams from Fashion MNIST or CIFAR-10, our metaplastic BNN outperforms a standard BNN and closely matches the accuracy of the network trained on the whole dataset. 
These results suggest that BNNs are more than a low precision version of full precision networks and highlight the benefits of the synergy between neuroscience and deep learning \cite{blake}. 
\vspace*{-0.6cm}
\subsection*{Hidden weights as metaplastic states}
\vspace*{-0.2cm}
The problem of forgetting in artificial neural networks results from a dilemma: synapses need to be updated in order to learn new tasks but also to be protected against further changes in order to preserve knowledge.
In a foundational neuroscience work, Fusi et al. show than in small Hopfield networks,
catastrophic forgetting can be addressed by introducing a hidden metaplastic state that controls the plasticity of the synapse \cite{fusi}.
Synapses can assume only $+1$ or $-1$ weight, with the metaplastic state modulating the difficulty for the synapse to switch.
Therefore, in this scheme, repeated potentiation of a positive-weight synapse will only affect its metaplastic state and not its actual weight.
Here, we remark that the way that BNNs are trained is remarkably similar to this situation. In BNNs, synapses can also only assume $+1$ or $-1$ weight, and they feature a hidden real weight ($W^{\rm h}$), which is updated by backpropagation. 
The synaptic weight changes between $+1$ and $-1$ only when $W^{\rm h}$ changes sign, suggesting that $W^{\rm h}$ can be seen as a metaplastic state modulating the difficulty for the actual weight to change sign.
However, standard BNNs are as prone to catastrophic forgetting as conventional neural networks.
In \cite{fusi}, Fusi et al. showed that the metaplastic changes should make subsequent affect plasticity exponentially to mitigate forgetting, whereas $W^{\rm h}$ affects weight changes only linearly in BNNs.
Therefore, in this work, we propose to adapt the learning process of BNNs  so that the larger the magnitude of a hidden weight $W^{\rm h}$, the more difficult to switch its associated binarized weight $W^{\rm b} = \mbox{sign}(W^{\rm h})$. 
Denoting $U_W$ the update provided by the learning algorithm, we implement:
\vspace{-0.2cm}
\begin{align*}
    W^{\rm h} &\leftarrow W^{\rm h} - \eta U_W \cdot f_{ \rm meta}(m, W^{\rm h}) \quad {\rm if} \quad U_W W^{\rm h}>0 \\
    W^{\rm h} &\leftarrow W^{\rm h} - \eta  U_W \quad {\rm otherwise.}
\end{align*}
As in the metaplasticity model of \cite{fusi} where synaptic plasticity decreases exponentially with the metaplastic state, we choose $f_{\rm meta}(m, W^{\rm h}) = {\rm tanh^{'}}(m\cdot W^{\rm h})$ to produce an exponential decay for large metaplastic states $W^{\rm h}$, where $m$ is an hyperparameter that controls the consolidation. 
\vspace{-0.6cm}
\subsection*{Toy problem study}
\begin{figure}[t]
\begin{center}
 \includegraphics[width=0.5\textwidth]{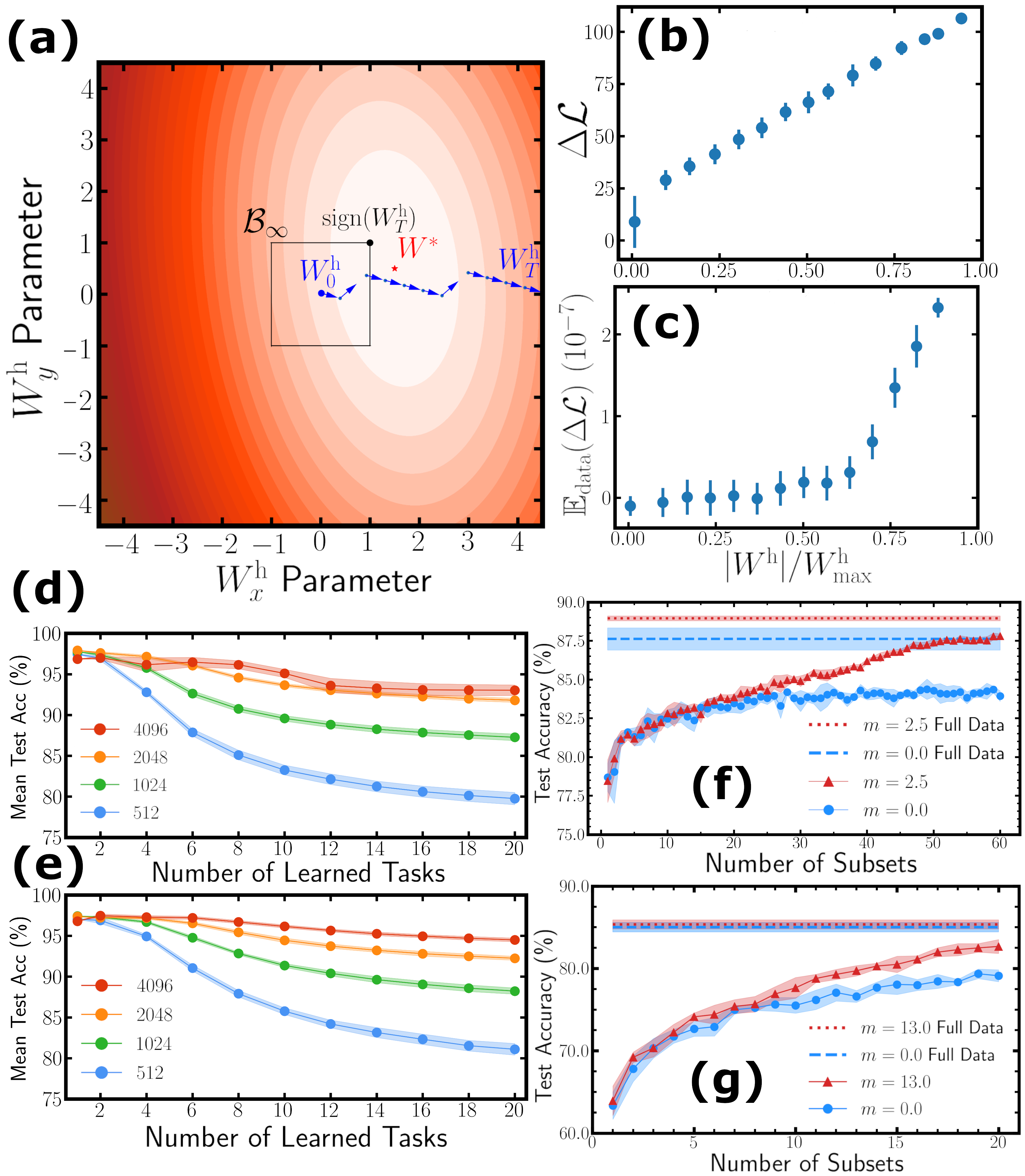}
\end{center}
  \caption{(a) Quadratic binarized optimization in two dimensions. (b-c) Average loss increase when switching $W^{\rm b}$ versus normalized hidden weight, for the binary quadratic problem (b) and for a BNN on MNIST (c). (d-e) Permuted MNIST benchmark with our method (d) and EWC (e), where the x axis labels the number of learned tasks, with one color per network size. Fashion MNIST (f) and CIFAR-10 (g) test accuracy in the stream learning setting allowed by our approach (red) compared to a standard BNN (blue). Horizontal rules denote full dataset training baseline.}
  \label{fig}
\end{figure}

\vspace*{-0.2cm}
To validate the interpretation of hidden weights as metaplastic states, we first focus on a highly simplified binary optimization task that we solve in a way analogous to the BNN training process.
We want the binarized weights $W^{\rm b}$ to minimize a quadratic loss $\mathcal{L}$, as depicted by the color map on Fig.~\ref{fig}(a) in two dimensions, with $W^*$ as the global optimum.
We assume that $W^{\rm h}$ is updated by loss gradients computed with binarized weights, similarly to BNNs:
\vspace{-0.4cm}
\begin{align*}
W^{{\rm h}}_{t+1} = W^{{\rm h}}_{t} - \eta \frac{\partial \mathcal{L}}{\partial W}(W^{\rm b}_{t}).
\end{align*}
We can show that if the infinite norm of $W^*$ is lesser than one, some hidden weights diverge as $t \to \infty$.
This is because $W^{\rm h}$ is updated by loss gradients computed at the corners of the square, in contrast with conventional optimization.
More importantly, if we define importance of the binarized weight as the increase of the loss $\Delta \mathcal{L}$ when the weight is switched to the opposite value, we can prove that the speed of divergence of the hidden weight is directly linked to the importance of the binarized weight. 
For instance, in Fig.~\ref{fig}(a), $W^{\rm b}_x$ is more important than $W^{\rm b}_y$ for optimization.
Finally, we plot $\Delta \mathcal{L}$ versus $|W^{\rm h}|$ in Fig.~\ref{fig}(b), (c) for higher dimensions and for a BNN trained on MNIST and observe that the correspondence between important weights and hidden weight divergence still holds, justifying the fact that consolidating synapses with diverging hidden weights as our proposal does, is a promising route for mitigating catastrophic forgetting.
\vspace*{-0.5cm}
\subsection*{Experimental results}
\vspace*{-0.2cm}
\paragraph{Continual learning benchmark.} We now apply our consolidation strategy to the permuted MNIST benchmark on two hidden layers perceptrons of varying number of neurons.
We show in Fig.~\ref{fig}(d),(e) the average test accuracy as a function of the number of tasks learned so far.
We observe that our technique indeed allows sequential task learning and performs almost as well as Elastic Weight Consolidation (EWC) \cite{kirkpatrick} adapted to BNNs (the importance factor is computed with the binarized weights) over a wide range of hidden layer sizes when learning up to 20 tasks.
We choose $m=1.35$ and $\lambda_{EWC}=5\cdot10^{-3}$ for EWC.
\vspace*{-0.6cm}
\paragraph{Learning from a stream of data.} By construction, our approach 
does not require to update the importance factor
between two consecutive tasks.
Building on this asset, we explore a new setting, which we call stream learning, and where a task is learned by learning sub-parts of the full dataset sequentially, with all classes evenly distributed in each subset.
We choose Fashion MNIST (FMNIST) and CIFAR-10 for our experiments.
The architectures used are a perceptron with two hidden layers of 1,024 units for FMNIST
and a VGG-16 convolutional architecture for CIFAR-10.
We plot on Fig.~\ref{fig}(f), (g) the test accuracy reached by those networks when metaplasticity is used (red) or not (blue). 
We see that our approach comes closer to the accuracy reached when the full dataset is learned at once (straight lines) than the non-metaplastic counterpart.
Overall, these results highlight the benefit of metaplasticity models from neuroscience when applied to machine learning.

\subsection*{Acknowledgement}
This work was supported by European Research Council Starting Grant NANOINFER (reference: 715872).

\end{document}